\documentclass{article}

\usepackage{microtype}
\usepackage{graphicx}
\usepackage{subfigure}
\usepackage{booktabs} % for professional tables
\usepackage{enumitem}
\usepackage{multirow}
\usepackage{amsfonts}

\usepackage{hyperref}
\usepackage{amsmath}
\usepackage[accepted]{icml2022}

\icmltitlerunning{Under review by the International Conference
on Machine Learning Workshop}

\begin{document}

\twocolumn[
\icmltitle{Learning Large-scale Universal User Representation with Sparse Mixture of Experts}



\icmlsetsymbol{equal}{*}

\begin{icmlauthorlist}
\icmlauthor{Caigao Jiang}{to}
\icmlauthor{Siqiao Xue}{to}
\icmlauthor{James Zhang}{to}
\icmlauthor{Lingyue Liu}{to}
\icmlauthor{Zhibo Zhu}{to}
\icmlauthor{Hongyan Hao}{to}

% \icmlauthor{Cieua Vvvvv}{goo}
% \icmlauthor{Iaesut Saoeu}{ed}
% \icmlauthor{Fiuea Rrrr}{to}
% \icmlauthor{Tateu H.~Yasehe}{ed,to,goo}
% \icmlauthor{Aaoeu Iasoh}{goo}
% \icmlauthor{Buiui Eueu}{ed}
% \icmlauthor{Aeuia Zzzz}{ed}
% \icmlauthor{Bieea C.~Yyyy}{to,goo}
% \icmlauthor{Teoau Xxxx}{ed}
% \icmlauthor{Eee Pppp}{ed}
\end{icmlauthorlist}

\icmlaffiliation{to}{Ant Group, Hangzhou, China}
% \icmlaffiliation{goo}{Googol ShallowMind, New London, Michigan, USA}
% \icmlaffiliation{ed}{School of Computation, University of Edenborrow, Edenborrow, United Kingdom}

\icmlcorrespondingauthor{Siqiao Xue}{siqiao.xsq@alibaba-inc.com}
\icmlcorrespondingauthor{James Zhang }{james.z@antgroup.com}

\icmlkeywords{Machine Learning, ICML}

\vskip 0.3in
]



\printAffiliationsAndNotice{}%\icmlEqualContribution} % otherwise use the standard text.

\begin{abstract}
Learning user sequence behaviour embedding is very sophisticated and challenging due to the complicated feature interactions over time and high dimensions of user features.  Recent emerging foundation models, \textit{e}.\textit{g}., BERT and its variants, encourage a large body of researchers to investigate in this field. However, unlike natural language processing (NLP) tasks, the parameters of user behaviour model come mostly from user embedding layer, which makes most existing works fail in training a universal user embedding of large scale.  Furthermore, user representations are learned from multiple downstream tasks, and the past research work do not address the seesaw phenomenon.  In this paper, we propose SUPERMOE, a generic framework to obtain high quality user representation from multiple tasks. Specifically, the user behaviour sequences are encoded by MoE transformer, and we can thus increase the model capacity to billions of parameters, or even to trillions of parameters. In order to deal with seesaw phenomenon when learning across multiple tasks, we design a new loss function with task indicators.  We perform extensive offline experiments on public datasets and online experiments on private real-world business scenarios. Our approach achieves the best performance over state-of-the-art models, and the results demonstrate the effectiveness of our framework.
\end{abstract}

\section{Introduction}
\label{Introduction}
% Traditionally, mining user behaviour patterns from mobile APPs logs rely on task specific and handcrafted features\cite{zeng2021knowledge}. These kinds of methods handle user features more effectively, however they are labour-consuming and maybe sub-optimal, which fail to enable more complex data representations. 

Recent works have demonstrated that the pre-trained model plays a critical role on a wide range of applications, \textit{e}.\textit{g}.,  \cite{devlin2018bert,dosovitskiy2020image,riquelme2021scaling,bommasani2021opportunities,geng2022recommendation,sun2019bert4rec,qiu2020pre,khan2021transformers,wu2020ptum,xiao2021uprec,zeng2021knowledge}.  To improve the efficiency and effectiveness of these models, many researchers attempt to exploit transformer in order to capture chronological pattern and dynamics of user intentions \cite{zeng2021knowledge,xue2021graphpp}.  With the remarkable achievements of pre-trained models, especially BERT-based models \cite{qiu2021u}, the transformer backbone has been utilized to address user data sparsity and cold-start problems in downstream applications \cite{yuan2020parameter,zhang2020general}.  In addition, DNN-based self-supervised learning (SSL) model is designed to improve semantic representations for highly-skewed data distribution, with inadequate explicit user feedback in user behaviour sequence interactions via unlabeled data \cite{yao2021self,shin2021scaling,zhang2020gcn}.

However, the existing pre-trained model suffers from many difficulties in achieving good user representations, \textit{e}.\textit{g}.,  only a few behaviour channels are used in the model due to the huge sizes of vocabularies and the resulting low training efficiency.  In AETN \cite{zhang2020general}, only three behaviour channels are utilized, yielding sub-optimal user representations.
% Different from NLP and CV applications, typically user behaviour data has its unique characteristics with highly sparse input space. 
Therefore, the motivations of our work are threefold, supported by our practical observations in online production system.  Firstly, most of model parameters come from feature embedding of ID and categorical features, which usually dominate GPU memory usage \cite{lian2021persia}.  For example, the number of user IDs are often in the scale of billions, resulting in parameter size of $\textit{numberIDs}\times\textit{embeddingDIMs}$.  Secondly, the front embedding layer accounts for the majority of the model's size, while the rest of model layers are extremely computationally expensive.  Consequently, training feature embedding layer and main neural networks simultaneously and synchronously for model of large scale is challenging, which calls for efficient model training algorithm for sparsity.  Finally, there are multiple training objectives no matter in model pre-training stage or in fine-tuning stage, which often causes pre-trained user embedding models with sub-optimal performance when using simple bottom-shared mechanism for the reason of seesaw phenomenon \cite{tang2020progressive}  and negative transfer \cite{ma2018modeling,chen2019behavior}.

\begin{table*}
  \label{tab:dataset}
  \begin{tabular}{c c c c c c}
    \toprule
    Model & Large channels&Sequential&Temporal&Multi-task learning& Scalability up to Trillions\\
    \midrule
    MTL\cite{tang2020progressive} & $\surd$ & $\times$&$\times$&$\surd$& $\times$\\
    PERSIA\cite{lian2021persia} & $\surd$& $\times$&$\times$&$\times$&$\surd$\\
    MTSSL\cite{yao2021self}& $\surd$& $\times$&$\times$&$\times$&$\times$\\
    BERT\cite{sun2019bert4rec} & $\times$& $\surd$&$\times$&$\surd$&$\surd$\\
    AETN\cite{zhang2020general} & $\times$ & $\surd$ &$\times$&$\surd$&$\times$\\
    OURS &$\surd$ &$\surd$&$\surd$&$\surd$&$\surd$\\
  \bottomrule
\end{tabular}
\caption{Advantages and limitations of the proposed model and the other models }
\label{tab:comparison}
\end{table*}
In this paper, we propose SUPERMOE, a general framework for user sequence behaviour representation and prediction using sparse MoE transformer.  Intuitively, transformer demonstrates the importance of capturing long range dependencies and pairwise or higher order interactions between elements \cite{bommasani2021}. The sparse gating mechanism, such as MoE, has shown its great advantages in multi-objective learning in user recommendation systems.  Therefore, embedding the gating function in transformer would be a good alternative to conventional models in user representation learning. The comparison of advantages and limitations of the proposed model and the other models is listed in Table ~\ref{tab:comparison}.  

Our contributions can thus be summarized as follows: \textbf{1)} We propose a sparse MoE transformer model to deal with huge amount of user behaviour sequence data with high dimensions. \textbf{2)} We propose a novel multi-task optimization algorithm in order to address seesaw problem and negative transfer problem across multiple tasks. \textbf{3)} We devise a novel method to split feature projection layer in order to address the issue of GPU memory explosion, which successfully integrates hundreds of behaviour channels into model training. \textbf{4)} Our method significantly outperforms existing user behaviour representation learning methods.

\section{Problem Statement}
Generally, we denote a typical one-channel user behaviour sequence as $s=[s_{1},s_{2},...,s_{i},...,s_{N}]$, where $s_{i}$ indicates the $i^{th}$ user behaviour for this channel, which has length of $N$. A multi-channel user behaviour sequence is denoted as $ S=\{[s^{j}_{1},s^{j}_{2},...,s^{j}_{i}...,s^{j}_{N}]\}$, and $[s^{j}_{1},s^{j}_{2},...,s^{j}_{i}...,s^{j}_{N}]$ is the $j^{th}$ channel of user behaviour sequence corresponding to $M$ behaviour channels. Each instance $S$ in each task contains a userID $u\in U$, and three types of sequence channels, namely, category channel $S_{category}$, ID channel $S_{ID}$ and dense channel $S_{dense}$. Therefore, given a set of $N$ tasks $T=\{t_{1}, t_{2},...,t_{n}\}$ with corresponding supervised label $Y = \{y_{1},y_{2},...,y_{n}\}$, our goal is to learn the base user representations across these tasks in order to apply them to downstream applications. Following the two-stage training paradigm \cite{devlin2018bert}, we pre-train a base model firstly on the huge pre-training dataset and then fine-tune a new model on downstream target dataset with parameters initialized as the pre-trained model. After the training, our base representation model should be able to produce universal representation $\mathcal{H}$ to serve all downstream tasks.

\section{Methodology}
% In this section, we present the framework of SUPERMOE, which consists of three stages and mainly utilizes MoE transformer to acquire high quality user embedding. Particularly, our general user embedding pre-training model is introduced in Section 3.1. And then in Section 3.2, we introduce our huge-vocab categorical multi-channel feature projection method to obtain fixed length low-dimensional embedding vector. 
% The overall architecture is shown in figure \ref{fig: arch}.
\begin{figure*}[ht]
  \centering
  \includegraphics[width=1.0\linewidth]{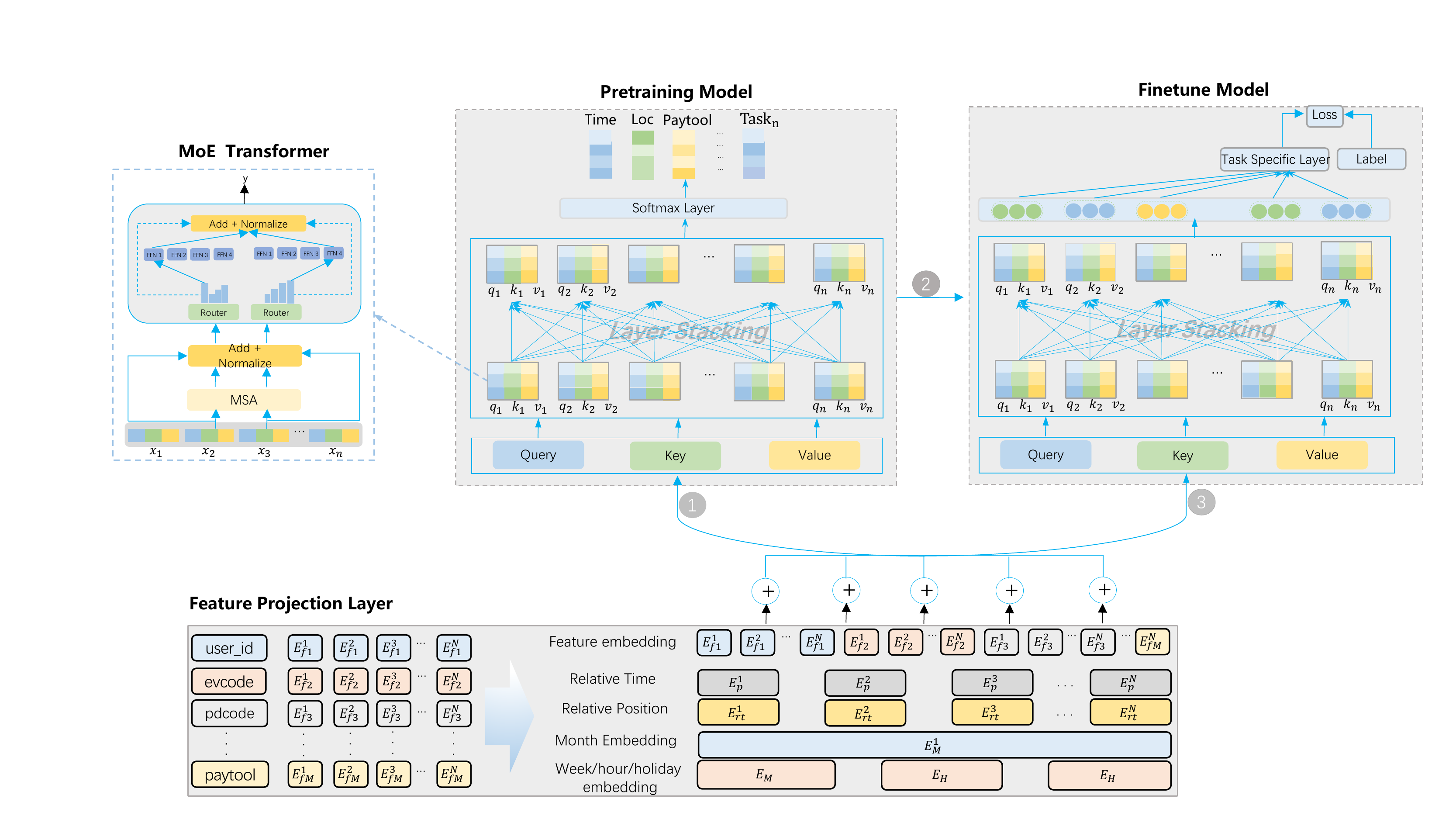}
  \caption{The \textbf{SuperMoE} framework consists of three different stages. In the multi-channel feature projection stage, all the channel features are embedded as dense vectors. During the pre-training stage, a series of masked channel prediction(MCP) tasks are utilized in order to achieve general user representations. The upper right shows the finetune stage, which freezes the parameters of pre-training model as an initialization. The upper left depicts a standard MoE transformer unit with dynamic routing mechanism.}
%   \Description{Average AUC Performance on Different Dataset}
  \label{fig: arch}
\end{figure*}

\subsection{User Embedding Pre-training Framework}

{\bfseries Pre-training Tasks.}
Similar to the pre-training task in \cite{devlin2018bert}, a new user representation pre-training task is designed to cater to the attribution of user behaviour data, i.e., \textit{masked channel prediction} (MCP) task.  Slightly different from \textit{masked language modeling} (MLM) task in NLP, not all of the features are masked due to multi-channel problem in user behaviour data which would produce too many feature vocabularies.  Theoretically, in the MCP task, some channel elements in the behaviour sequence are randomly masked with special token $[MASK]$ at pre-training stage. Therefore, an MCP task of one feature channel is elaborated as $input=[s_{1},s_{2},...,[MASK]_{i}...,s_{N}]$, with $ label=[MASK]_{i}$.  However, only a few channels are selected to be MCP tasks due to our belief that the more important a user behaviour sequence is, the more likely the sequence is selected as MCP task. In order to preserve essential information of user behaviours, we choose user ID, location, time interval, payment tool, product, trade amount, super position model (SPM) trace, click, and conversion etc.

{\bfseries Pre-training Objectives.} 
Formally, we denote $s_{mask}$ as the probability of the estimated activity, and the probability $p(s_{mask};\Theta)$ is represented by the product of the conditional distributions over the masked sequence:
\begin{equation}
    p(s_{mask};\Theta) = \prod_{i=1}^{N} p(s_{mask}|s_{1},s_{2},...,[MASK]...,s_{N};\Theta)
\end{equation}
Our objective is to maximize $p(s_{mask};\Theta)$, which is equivalent to minimizing the following loss function:
\begin{equation}
L_{mcp}^{i} = -\frac{1}{|S_{i}|}\sum_{j\in S_{i}}-log p(\hat{s_{j}}=s_{j})\textrm{ , }
\end{equation}
where $S_{i}$ is the set of positions of masked elements of the $i^{th}$ MCP task, and $\hat{s_{j}}$ and $s_{j}$ are the predicted user behaviour and the ground-truth behaviour, respectively.  Notably, user behaviours are of very different statistical characteristics from NLP or CV, e.g., the click and conversion task are sequential tasks. Hence, we propose a new training objective function:
\begin{equation}
L_{k}(\Theta_{k}) = \frac{1}{\sum_{i}{\delta_{k}^{i}}}\sum_{i}{\delta_{k}^{i}loss_{k}(\widehat{s}_{k}^{i}(\Theta_{k}),s_{k}^{i})}\textrm{ , }
\end{equation}
where $\delta$ is the indicator of training samples among $k$ tasks.

{\bfseries Pre-training Model Framework.} The main architecture ingredients of pre-training model are a stack of MoE transformers. Basically, our MoE transformer's backbone has a simple structure which consists of a multi-channel feature projection (MFP) layer, a MoE multi-head self-attention (MoE-MSA) layer and two MoE feed-forward network (MoE-FFN) layers. MFP layer takes the following form:
\begin{equation}
     y_{mpf}= [s_{category}*w_{category},\ split(s_{ID}*w_{ID}),\ s_{dense}]\textrm{ , }
\end{equation}
where $[\cdot]$ means the concatenation operator of all vectors.  Each MFP layer in the encoder block is followed by a layer normalization and nonlinear activation layer. The operator $split(\cdot)$ is a model parallel operation, implemented by the whale framework \cite{280776}.  Note that the splitting of MFP layer addresses the issue of GPU memory explosion, which successfully integrate hundreds of behaviour channels into model training.  An MoE-MSA layer takes the output of MFP $y_mpf$ as input, formulated as:
\begin{equation}
y_{msa}=softmax(\frac{(qw^{q}G_{q}(q))(kw^{k}G_{k}(k))^{T}}{\sqrt{d_{k}}})(vw^{v}G_{v}(v))\textrm{ , }
\end{equation}
% Where $G_{Q}(Q)=Softmax(TopK(H_{e}(Q), k))$, $G_{K}(K)=Softmax(TopK(H_{e}(K), k))$, and $G_{V}(V)=Softmax(TopK(H_{e}(V), k))$.
where $q,k,v$ is the output of an MFP layer, and $y_{msa}$ is output of an MoE-MSA layer, connected by two MoE-FNN layers.  Lastly, the point-wise MoE-FNN\cite{fedus2021switch} can be formulated as:
\begin{equation}
y_{ffn}=\sum_{e=1}^{E}G_{e}(x)\cdot \textit{FFN}_{e}(x)\textrm{ , }
\end{equation}
with $\textit{FFN}_{e}(x)=w_{o_{e}}\cdot \textit{Relu}(w_{i_{e}}\cdot x)$,$G_{e}(x)=softmax(\textit{TopK}(h_{e}(x), k))$ , where $w_{o}$ and $w_{i}$ are the standard feed-forward networks with the same parameters.  We choose top 1 strategy \cite{fedus2021switch} for $\textit{TopK}(\cdot)$ function.  In summary, $y_{ffn}$ is the output of backbone of an MoE transformer.  Formally, a series of MoE transformer blocks can be described as:
\begin{equation}
y_{moe} = \textit{MoETransformer}([s_{category},s_{ID},s_{dense}])\textrm{ , }
\end{equation}
where $\textit{MoETransformer}=$$\textit{MoE}_{FFN}(\textit{MoE}_{MSA}(\textit{MFP}(\cdot)))$.  The overall pre-training architecture is shown in Figure \ref{fig: arch}.

\subsection{User Embedding Fine-tuning Framework}
After pre-training, we adapt the learned user representations to specific downstream tasks, instead of using pre-trained representations directly, which is somehow unrelated to our defacto targets in production environment. Therefore, we need to develop a new model to fine-tune our user behaviour model across multiple downstream tasks with a unified framework. Assuming that we have restored and initialized the parameters of the previous pre-trained model, the fine-tuning model shares the parameters of the pre-trained part, and a linear classification layer is placed on the top of the final output without activation function.  Denoting $h_{o}$ as the output of the final MoE-transformer, we have:
\begin{equation}
y_{i} = \textit{Tower}_{i}(\textit{MaxPooling}(h_{o}))\textrm{ , }
\end{equation}
and the
% $y_{i} = MoETransformer_{1~N}([s_{category},s_{ID},s_{dense}])$ and 
$Tower_{i}$ is a linear classification layer of the $i^{th}$ fine-tuning task. Note that the user representation $\mathcal{H}=\textit{MaxPooling}(h_{o})$.  The overall architecture of our fine-tuning framework is shown in Figure \ref{fig: arch}. 
\subsection{Multi-task Training Optimization}
% In this section, we introduce our hybrid distributed training algorithm and pre-training tasks. For the sake of understanding, the fine-tuning tasks is presented in experiment part.
In order to address seesaw and negative transfer problems and to improve learning from multiple tasks, such as regression and classification, we leverage a multi-task optimization strategy, i.e., jointly optimize across multiple tasks, which can be applied in both pre-training stage and fine-tuning stage. Mathematically, we get $k$ training objectives from equation ($5$), and therefore, the total loss can be formulated as:
\begin{equation}
\begin{split} 
    & Loss(\Theta) = \lambda_{1} *l_{1}(\widehat{s}_{1}(\Theta_{1}),s_{1})+
    \\
    &\lambda_{2} *l_{2}(\widehat{s}_{2}(\Theta_{2}),s_{})+...+\lambda_{k} *l_{k}(\widehat{s}_{k}(\Theta_{k}),s_{k})\textrm{ , }
\end{split}
\end{equation}
where $Loss(\Theta)$ denotes the total loss and $\alpha_{k}$ is the regularization strength of the $k^{th}$ loss. Recall that our objective is actually to maximize Area Under Curve (AUC) score, we consider the following bi-level optimization problem:
\begin{equation}
Max\ AUC_{val}(\theta_{\lambda},\lambda)\ \ ~~s.t. \theta_{\lambda}=\arg\min_{\Theta}Loss(\Theta,\lambda)\textrm{ , }
\end{equation}
where $AUC_{val}$ is the AUC score on validation dataset while training.  However, $AUC_{val}(\theta_{\lambda},\lambda)$ is non-differentiable with the indicator function $\mathbb{I}(f(\lambda,x_i^+)<f(\lambda,x_j^{-}))$, and $x_i^+$ and $x_j^{-}$ are the positive and negative samples, respectively. We therefore employ $max\{0,1-(f(\lambda,x_i^+)-f(\lambda,x_{j}^{-}))\}$ as a differentiable convex surrogate of the above indicator function.
% \subsubsection{\textbf{Training Tasks}}~\\
% \textbf{Masked Channel Prediction(MCP):}
% This task is the main pre-training task, which is similar to the masked language model task in BERT. As we stated, for the large number of channels and vocabularies of the user behaviours, therefore, we address this problem with selected 11 channels. We calculate the sum of all MCP task losses by averaging the softmax output of cross entropy function, and we denote it as $L_{mcp}$.
% ~\\
% \textbf{Next Time Prediction(NTP):}
% This task is to predict the next time when user's payment occurs, which is a regression problem. A mean square error(MSE) loss function is utilized, and denoted as $L_{ret}$.
% ~\\
% \textbf{Click-Through-Conversion Rate(CTCVR) Task:}
% To discovery users' preferences for products from a vast number of candidates, generally, we use Click-Through-Conversion Rate(CTCVR) to estimate the probability of an product from impression to click and from click to purchase respectively. Unlike MCP task, the CTCVR task is a cascading behaviour with sample selection bias and data sparsity problem.

% \subsubsection{\textbf{Task-Related Multi-task Training Process}}
\section{Experimental Methodology}
In this section, we demonstrate the online and offline performance of SUPERMOE in generating general embedding for user behaviour sequence. We evaluate our model in four different real world test datasets, and one for public and three for private datasets respectively. 
% We compare our model with four different baseline models in the downstream tasks.
\begin{table*}
  \caption{Overall AUC performance for different models}
  \label{tab:res_overall}
  \begin{tabular}{ccccl}
    \toprule
    Model &SIUPD &PAYTOOL &MCP&\qquad FORTUNE\\
    & Category1\quad Category2 & Category1\quad Category2\quad Category3\quad Category4\quad Category5& subscription & CTR\qquad CVR\\
    \midrule
    MMOE & 80.758 \quad\qquad 79.172&87.691 \qquad\ 55.016 \quad\qquad92.166 \quad\qquad61.019 \qquad90.581&67.843 & 80.988\quad90.765\\
    PLE& 81.819 \quad\qquad 79.798&87.762 \qquad\ 55.269 \quad\qquad92.803 \quad\qquad61.267 \qquad91.924&68.351 & 81.719\quad91.751\\
    BERT& 83.021 \quad\qquad 80.319&88.908 \qquad\ 56.081 \quad\qquad93.217 \quad\qquad62.832 \qquad93.657&70.092& 82.683\quad92.014\\
    AETN&82.828 \quad\qquad 80.774&89.293 \qquad\ 55.961 \quad\qquad93.229 \quad\qquad62.706 \qquad92.899&70.055& 82.952\quad91.817\\
    \textbf{OURS}& \textbf{83.453} \quad\qquad \textbf{80.971}&\textbf{89.598} \qquad\textbf{56.192}\quad\qquad\textbf{93.461} \quad\qquad\textbf{63.574}\qquad\textbf{94.356}&\textbf{71.218} &\textbf{83.791}\quad\textbf{92.331}\\
    \bottomrule
  \end{tabular}
\end{table*}
\subsection{Experiment Settings}
\subsubsection{\textbf{Dataset description}}
We evaluate the performance of our model on four different downstream applications, i.e., SIUPD, Paytool, MCP, and Fortune.  SIUPD dataset comes from the IJCAI17 contest \footnote{https://tianchi.aliyun.com/dataset/dataDetail?dataId=58}, which contains 139,6245 users' shopping logs on Alipay platform.  Paytool is a user payment preference dataset, which describes the behaviour of using payment tools for online users. In MCP dataset, we use 103 channels of subscription and redemption behaviour sequences for users. Fortune dataset includes users "impression$\rightarrow$click" and "click$\rightarrow$purchase" behaviours.  All these four datasets are split into training/test sets with the ratio of $0.8$/$0.2$.  The statistics of the datasets can be found in 
Table ~\ref{tab:dataset}.
\begin{table}[h]
\caption{Dataset Descriptions}
  \label{tab:dataset}
  \begin{tabular}{c c c c c}
    \toprule
    Dataset & Training&Test&Channels&AverageLength\\
    \midrule
    SIUPD & 16M& 4M&11& 150\\
    Paytool & 240M& 60M&12& 128\\
    MCP & 80M &20M &103& 128\\
    Fortune & 32M& 8M&786& 128\\
  \bottomrule
\end{tabular}
\end{table}

\subsubsection{\textbf{Baselines}}
We fine-tune and evaluate our model against four other representative models: \textbf{MMOE\cite{ma2018modeling}}, a classical multi-task recommendation model, \textbf{PLE\cite{tang2020progressive}}, an extension of MMOE with multiple progressive extraction layers, \textbf{BERT\cite{devlin2018bert}}, a well-famed sequence model widely used in large scale representation learning, especially in NLP and \textbf{AETN\cite{zhang2020general}}, a user representation learning model, which combines multi-head attention and Denoising Autoencoder(DAE) model to generate user embeddings.

% \begin{table*}
%   \caption{Overall AUC performance for different models}
%   \label{tab:res_overall}
%   \begin{tabular}{ccccl}
%     \toprule
%     Model &SIUPD &PAYTOOL &MCP&\qquad FORTUNE\\
%     & Category1\quad Category2 & Category1\quad Category2\quad Category3\quad Category4\quad Category5& subscription & CTR\qquad CVR\\
%     \midrule
%     MMOE & 80.758 \quad\qquad 79.172&87.691 \qquad\ 55.016 \quad\qquad92.166 \quad\qquad61.019 \qquad90.581&67.843 & 80.988\quad90.765\\
%     PLE& 81.819 \quad\qquad 79.798&87.762 \qquad\ 55.269 \quad\qquad92.803 \quad\qquad61.267 \qquad91.924&68.351 & 81.719\quad91.751\\
%     BERT& 83.021 \quad\qquad 80.319&88.908 \qquad\ 56.081 \quad\qquad93.217 \quad\qquad62.832 \qquad93.657&70.092& 82.683\quad92.014\\
%     AETN&82.828 \quad\qquad 80.774&89.293 \qquad\ 55.961 \quad\qquad93.229 \quad\qquad62.706 \qquad92.899&70.055& 82.952\quad91.817\\
%     \textbf{OURS}& \textbf{83.453} \quad\qquad \textbf{80.971}&\textbf{89.598} \qquad\textbf{56.192}\quad\qquad\textbf{93.461} \quad\qquad\textbf{63.574}\qquad\textbf{94.356}&\textbf{71.218} &\textbf{83.791}\quad\textbf{92.331}\\
%     \bottomrule
%   \end{tabular}
% \end{table*}

% \subsubsection{\textbf{Evaluation Metrics}}
% We adopt the widely used Area Under Curve(AUC), precision and recall as our metrics. Specifically, the precision and recall as the metrics for online evaluation. As for offline evaluation, we also report AUC scores for measuring the offline performance.

\subsection{Offline Evaluation Results}
In order to show the advantages of our model, we conduct the following intrinsic experiments to evaluate offline and online performances. 
% \begin{table*}
%   \caption{Overall AUC performance for different models}
%   \label{tab:res_overall}
%   \begin{tabular}{ccccl}
%     \toprule
%     Model &SIUPD &PAYTOOL &MCP&\qquad FORTUNE\\
%     & Category1\quad Category2 & Category1\quad Category2\quad Category3\quad Category4\quad Category5& subscription & CTR\qquad CVR\\
%     \midrule
%     MMOE & 80.758 \quad\qquad 79.172&87.691 \qquad\ 55.016 \quad\qquad92.166 \quad\qquad61.019 \qquad90.581&67.843 & 80.988\quad90.765\\
%     PLE& 81.819 \quad\qquad 79.798&87.762 \qquad\ 55.269 \quad\qquad92.803 \quad\qquad61.267 \qquad91.924&68.351 & 81.719\quad91.751\\
%     BERT& 83.021 \quad\qquad 80.319&88.908 \qquad\ 56.081 \quad\qquad93.217 \quad\qquad62.832 \qquad93.657&70.092& 82.683\quad92.014\\
%     AETN&82.828 \quad\qquad 80.774&89.293 \qquad\ 55.961 \quad\qquad93.229 \quad\qquad62.706 \qquad92.899&70.055& 82.952\quad91.817\\
%     \textbf{OURS}& \textbf{83.453} \quad\qquad \textbf{80.971}&\textbf{89.598} \qquad\textbf{56.192}\quad\qquad\textbf{93.461} \quad\qquad\textbf{63.574}\qquad\textbf{94.356}&\textbf{71.218} &\textbf{83.791}\quad\textbf{92.331}\\
%     \bottomrule
%   \end{tabular}
% \end{table*}

\subsubsection{\textbf{Offline Model Performance}}
In this section, we present the results of offline model performance in the downstream tasks. Table ~\ref{tab:res_overall} summarizes the overall AUC scores of different models across all datasets. Taking the evaluation results of SIUPD dataset as an example, it is obvious that our model improves the baseline method MMoE by gains of 2.7 and 1.8, respectively, in two combined tasks, for the reason that our model utilizes more abundant chronological user behaviours to address the behaviour sparsity issue. Moreover, we outperform the other two sequential models with gains of 0.63 and 0.19, respectively, benefiting from of our multi-task optimization. Similar performances can be observed in other three datasets.  It is worth mentioning that our methods all achieve the state-of-the-art performances with significant gains.
% \begin{figure}[h]
%   \centering
%   \includegraphics[width=1.1\linewidth]{averge_auc.pdf}
%   \caption{Average AUC Performance on Different Dataset}
%   \Description{Average AUC Performance on Different Dataset}
%   \label{overall_auc}
% \end{figure}

\subsubsection{\textbf{Offline Embedding Performance}}
To evaluate the user embedding quality and efficiency of our model, we conduct six different experiments for comparison, and analyze the effects of different embedding methods, as well as different model capacities. We select the user's payment switching task in PAYTOOL dataset to report AUC score, Recall@85 and Recall@50 respectively. The results are illustrated in Table \ref{tab:paytool}. Notably, all sequential embedding methods are better than PLE-only model, which demonstrates the advantage of user embedding.  Furthermore, our embedding is more effective than other two sequential models, which takes the same model size of 1 billion.  We also investigate the performance of different model capacities, and it can be seen in Table \ref{tab:paytool} that MoE with 20 billions parameters performs much better than MoE with 1 billion, which generates gains of 0.67 AUC, 2.01 recall@85, and 1.39 recall@50, respectively.
\begin{table}
\caption{Embedding Evaluation in PAYTOOL}
  \label{tab:paytool}
  \begin{tabular}{c c c c c}
    \toprule
    Model&AUC Score&Recall@85&Recall@50\\
    \midrule
    PLE &92.183 & 19.583&46.581\\
    PLE+BERT & 94.067&28.751& 50.673\\
    PLE+AETN & 94.143 &29.033&50.894\\
    PLE+MoE1B & 95.721&30.628& 53.766\\
    PLE+MoE10B & 96.169&31.193& 54.938\\
    \textbf{PLE+MoE20B} &\textbf{96.395} &\textbf{32.640}& \textbf{55.174}\\
  \bottomrule
\end{tabular}
\end{table}

\subsubsection{\textbf{Online A/B Testing}}

To further investigate the quality and effectiveness of our user embeddings, we conduct two A/B testing experiments against online baseline model. "Online1" experiment is a payment switching scenario operating on real-world Alipay platform. In this experiment, our model brings on gains of 13.41\% pv, 1.97\% in conversion and 21.36\% GMV.  In addition, our model achieves gains of 4.95\%,9.11\% and 25.19\%, respectively, in "Online2" experiment, which is a fund subscription and redemption scenario.
These results are summarized in Table ~\ref{tab:online}.

\begin{table}
 \caption{Online Comparison of Different Models}
  \label{tab:online}
  \begin{tabular}{cccccc}
    \toprule
    Scenario&Models&PV&PVCVR&GMV\\
    \midrule
    \multirow{2}{*}{Online1} & PLE+BERT & 0& 0& 0\\
                               & \textbf{OURS}&13.41\%&1.97\%&21.36\%\\
     \hline
    \multirow{2}{*}{Online2} &PLE+BERT &0&0&0 \\
                               &\textbf{OURS}&4.95\%&9.11\%&25.19\%\\
  \bottomrule
\end{tabular}
\end{table}

\section{Conclusions}
% In this paper, we investigated the utilization of multi-layer MoE networks as a practical way to massively increase model capacity.
% We have shown that it is possible to learn large scale user embeddings, while capturing ubiquitous high order correlations using sparse MoE, with our meticulous model architecture. Extensive empirical experiments demonstrated the overwhelming superiority of our method on various real-world datasets comparing to other state-of-the-art methods.

In this paper, we investigated the utilization of multi-layer MoE networks as a practical way to massively increase model capacity and to deal with seesaw phenomenon and negative transfer problem. To complete this research, we introduce an user behaviour representation pre-training and fine-tuning model using sparse MoE. We have shown that it is possible to learn large scale user embeddings, while capturing ubiquitous high order correlations using sparse MoE, with our meticulous model architecture. Moreover, we formulated a bi-level optimization method in order to address multi-task optimization. Extensive empirical experiments demonstrated the overwhelming superiority of our method on various real-world datasets comparing to other state-of-the-art methods.

\nocite{langley00}

\bibliography{main}
\bibliographystyle{icml2021}

% \clearpage
% \appendix
% \appendixpage

% \section{Dataset details}
% The details of the datasets used in the experiments can be found in Table \ref{tab:dataset}.

%%%%%%%%%%%%%%%%%%%%%%%%%%%%%%%%%%%%%%%%%%%%%%%%%%%%%%%%%%%%%%%%%%%%
%%%%%%%%%%%%%%%%%%%%%%%%%%%%%%%%%%%%%%%%%%%%%%%%%%%%%%%%%%%%%%%%%%%%%%%%%%%%%%%

\end{document}